%% file: main.tex
\documentclass{article}

\usepackage[preprint]{corl_2026} 

\usepackage[utf8]{inputenc} 
\usepackage[T1]{fontenc}    
\usepackage{hyperref}       
\usepackage{url}            
\usepackage{booktabs}       
\usepackage{amsfonts}       
\usepackage{nicefrac}       
\usepackage{microtype}      
\usepackage{graphicx}
\usepackage{subcaption}
\usepackage{caption}
\usepackage{capt-of}
\usepackage[textsize=tiny]{todonotes}
\usepackage[most]{tcolorbox}
\usepackage{enumitem}
\usepackage{xspace,soul}
\usepackage{duckuments}
\usepackage{lipsum}
\usepackage{multirow}
\input{color_utils}

\usepackage{algorithm}
\usepackage[noend]{algpseudocode}

\algrenewcommand{\algorithmiccomment}[1]{\hfill \xslategray{\texttt{\#~#1}}}
\usepackage[most]{tcolorbox}
\usepackage{wrapfig}
\usepackage{booktabs}

\usepackage{fvextra}
\usepackage[table]{xcolor}
\usepackage{enumitem}
\usepackage{placeins}
\usepackage{tabularx}

\newcommand{\hc}{\cellcolor{xblue!8}}

\usepackage[most]{tcolorbox}
\tcbuselibrary{listings,breakable,skins}

\newtcblisting{promptbox}[2][]{
  title={#2},
  colback=gray!5,
  colframe=gray!50!black,
  fonttitle=\bfseries\footnotesize,
  arc=2mm,
  boxrule=0.5pt,
  left=5pt, right=5pt, top=4pt, bottom=4pt,
  breakable,
  listing only,
  listing options={
    basicstyle=\ttfamily\scriptsize,
    breaklines=true,
    columns=fullflexible,
    keepspaces=true
  },
  #1
}

\newcommand{\evis}{\bar{e}_{\mathrm{vis}}}
\newcommand{\etext}{\bar{e}_{\mathrm{text}}}

\title{FRAME: Factored Retrieval via Attribute Readouts \\
    for Object-Centric Scene Memory}

\author{
  Woosang Jeon$^*$\\
  Seoul National University\\
  \texttt{jwoosang1@snu.ac.kr}  
  \And
  Sanghyeok Choi$^*$\\
  Seoul National University\\
  \texttt{cholsang83@snu.ac.kr}
  \AND
  Minwoo Kim\\
  Seoul National University\\
  \texttt{minwoo.kim@snu.ac.kr}
  \And
  Taehyun Jung\\
  Seoul National University\\
  \texttt{jth135@snu.ac.kr}
  \And
  Taehyeong Kim$^\dagger$\\
  Seoul National University\\
  \texttt{taehyeong.kim@snu.ac.kr}
}

\begin{document}
\maketitle

\begingroup
\renewcommand\thefootnote{}
\footnotetext{$^*$Equal contribution. \quad $^\dagger$Corresponding author.}
\endgroup

\begin{abstract}
Language-guided robots need persistent scene memories to follow instructions, revisit objects, and resolve references to objects encountered over time.
While much of language-guided scene-memory retrieval has emphasized spatial or relational references, many everyday object references specify objects by multiple persistent attributes, such as category, material, size, or surface appearance.
We formalize this problem as \emph{attribute-compositional retrieval}, where a fixed object-centric scene memory is queried with natural language to retrieve the object satisfying the requested attributes.
To investigate this capability directly, we introduce a controlled evaluation protocol with fixed scene memories and attribute-defined targets, separating retrieval from perception and annotation ambiguities.
We then propose FRAME, which turns language into query-relevant attribute weights, uses learned readouts to estimate per-attribute evidence from object embeddings, and ranks objects by aggregating this evidence according to the query.
Across held-out scenes and object assets, FRAME outperforms representative scene-memory retrieval baselines while reducing post-decomposition object scoring to lightweight matrix-vector computation.
These results position attribute-compositional retrieval as a complementary scene-memory capability for language-guided robots, showing that persistent object attributes can be exposed as composable evidence for accurate and efficient multi-attribute retrieval.
\end{abstract}




\section{Introduction}
\label{sec:intro}

Language-guided robots need persistent scene memories to follow instructions, revisit objects, and resolve references to objects encountered over time~\citep{kim20193,gu2024conceptgraphs,werby2024hierarchical}.
A common abstraction for such memories is an object-centric scene graph, in which each node stores visual and semantic representations of an object (e.g., appearance features, visual-language embeddings, or semantic labels), while edges encode spatial or geometric relations between objects~\citep{rosinol20203d, gu2024conceptgraphs}.
This structure naturally supports spatial references such as \textit{``the chair next to the table''} or \textit{``the mug on the shelf''}, and much of language grounding in 3D scenes has therefore focused on spatial anchors, relational expressions, and graph-based reasoning over object relations~\citep{chen2020scanrefer,achlioptas2020referit3d,zhang2023multi3drefer}.

However, many everyday references identify objects through conjunctions of persistent attributes rather than relations to other objects, such as category, material, size, and surface appearance.
For example, given the query ``the wooden chair,'' the correct object is the \textit{chair} made of \textit{wood}, whereas both a wooden table and a plastic chair should be excluded.
This raises a key question about retrieval: given an object-centric scene graph used as scene memory, how should language queries involving multiple object attributes be resolved?
We study this problem as \emph{attribute-compositional retrieval} and investigate it with a controlled evaluation protocol that fixes the scene memory and defines retrieval targets based on predefined attribute combinations.

Under this setting, common retrieval approaches exhibit complementary limitations.
Embedding-based approaches typically compare the language query against each object embedding with a single query--object score, making it difficult to distinguish full attribute matches from partial ones.
LLM/VLM-mediated and graph-based approaches can make attribute reasoning more explicit, but iterative query-time reasoning over scene context or candidate objects is computationally costly and ties object-attribute estimation to each query-time reasoning call~\citep{gu2024conceptgraphs,linok2025beyond,werby2024hierarchical,booker2024embodiedrag}.
We therefore propose FRAME---\textbf{F}actored \textbf{R}etrieval via \textbf{A}ttribute Readouts for Object-Centric Scene \textbf{ME}mory.
FRAME reads out query-relevant attribute evidence from object embeddings in a composable form, allowing multi-attribute references to be matched by combining the corresponding readout scores.
In held-out scenes and object assets, FRAME outperforms representative scene-memory retrieval baselines on this task while reducing post-decomposition object scoring to lightweight matrix-vector computation.

\paragraph{Contributions.}
(1) We formulate attribute-compositional retrieval over object-centric scene graphs: retrieving the object that satisfies a query-specified set of persistent attributes.
(2) We introduce a controlled evaluation protocol for attribute-compositional retrieval that uses fixed scene memories and defines retrieval targets based on predefined combinations of object attributes, isolating retrieval performance from perception and annotation ambiguity.
(3) We propose FRAME, an attribute-compositional retrieval method that decomposes language into attribute primitives, reads out attribute-specific evidence from object embeddings, and combines the resulting evidence according to the query to score candidate objects.


\section{Related Work}
\label{sec:related}

\paragraph{Benchmarks for language grounding in 3D scenes.}
Datasets such as ScanRefer~\citep{chen2020scanrefer}, ReferIt3D~\citep{achlioptas2020referit3d}, Multi3DRefer~\citep{zhang2023multi3drefer}, Scan2Cap~\citep{chen2021scan2cap}, MMScan~\citep{lyu2024mmscan}, and IRef-VLA~\citep{zhang2025iref} have established 3D grounding as selecting target object(s) from language in a 3D scene.
They are especially valuable for evaluating spatial, relational, viewpoint-dependent, and context-rich grounding.
Although such benchmarks contain attribute-like descriptions, these cues are usually entangled with category, spatial anchors, relations, viewpoint language, and broader scene context.
As a result, they do not evaluate attribute composition in isolation.
We instead study composition of persistent attributes directly as attribute-compositional retrieval over a fixed scene memory.

\paragraph{3D scene graphs and scene memory.}
Persistent 3D scene representations provide the substrate for language-guided embodied agents.
Prior work has studied 3D scene graphs for semantic mapping, actionable spatial perception, indoor scene-graph prediction, and semantic object-goal inspection~\citep{kim20193,rosinol20203d,rosinol2021kimera,wald2020learning,ginting2024seek}.
Recent open-vocabulary and hierarchical scene-memory systems, including ConceptGraphs~\citep{gu2024conceptgraphs}, HOV-SG~\citep{werby2024hierarchical}, OpenScene~\citep{peng2023openscene}, and OpenMask3D~\citep{takmaz2023openmask3d}, build rich object-centric memories from visual, textual, and geometric evidence.
Other memory-centric systems extend this direction toward long-horizon robot memory, episodic recall, or viewpoint-based observations, such as ReMEmbR~\citep{anwar2025remembr}, 3D-Mem~\citep{yang20253d}, and Mind Palace~\citep{ginting2025enter}.
Together, these works make persistent scene memory increasingly available to robots, but leave open how such memories should be queried once constructed.

\paragraph{LLM/VLM-mediated scene-memory querying.}
Recent approaches use LLMs or VLMs at inference time to mediate between natural-language queries and 3D scene memory.
Some rank objects from graph context, captions, and spatial relations~\citep{gu2024conceptgraphs,linok2025beyond}; others retrieve candidates or decompose queries before LLM/tool-based reranking or grounding~\citep{werby2024hierarchical,booker2024embodiedrag,yang2024llm}.
These approaches are flexible for open-ended, relational, and task-level language, but their flexibility comes with a cost.
For persistent attribute-centric retrieval, whether an object satisfies the queried attributes is determined at inference time rather than represented explicitly as reusable, query-independent object-side evidence.
As a result, they typically require iterative query-time reasoning over scene context or candidate objects.

\paragraph{Attribute recognition and compositionality.}
Attribute detection and compositional recognition have been extensively studied in 2D and multimodal representation learning, including open-vocabulary attribute detection~\citep{bravo2023open} and analyses of compositional behavior in vision-language representations~\citep{ahn2025can}. 
Open-vocabulary 3D understanding benchmarks, including OpenScan~\citep{zhao2026openscan}, also evaluate attribute-related recognition in 3D scenes.
Nevertheless, recent work suggests that compositional retrieval remains challenging even when attribute--object associations are encoded in vision-language embeddings~\citep{koishigarina2025clip}.
We study this gap in the context of persistent scene memory, treating attributes as composable object-side evidence for language-guided retrieval.


\section{Attribute-Compositional Retrieval Protocol}
\label{sec:protocol}

\begin{figure}[t]
\centering
\vspace{-1.5em}
\includegraphics[width=0.98\linewidth,trim={0.5cm 0.0cm 0.5cm 0.5cm}]{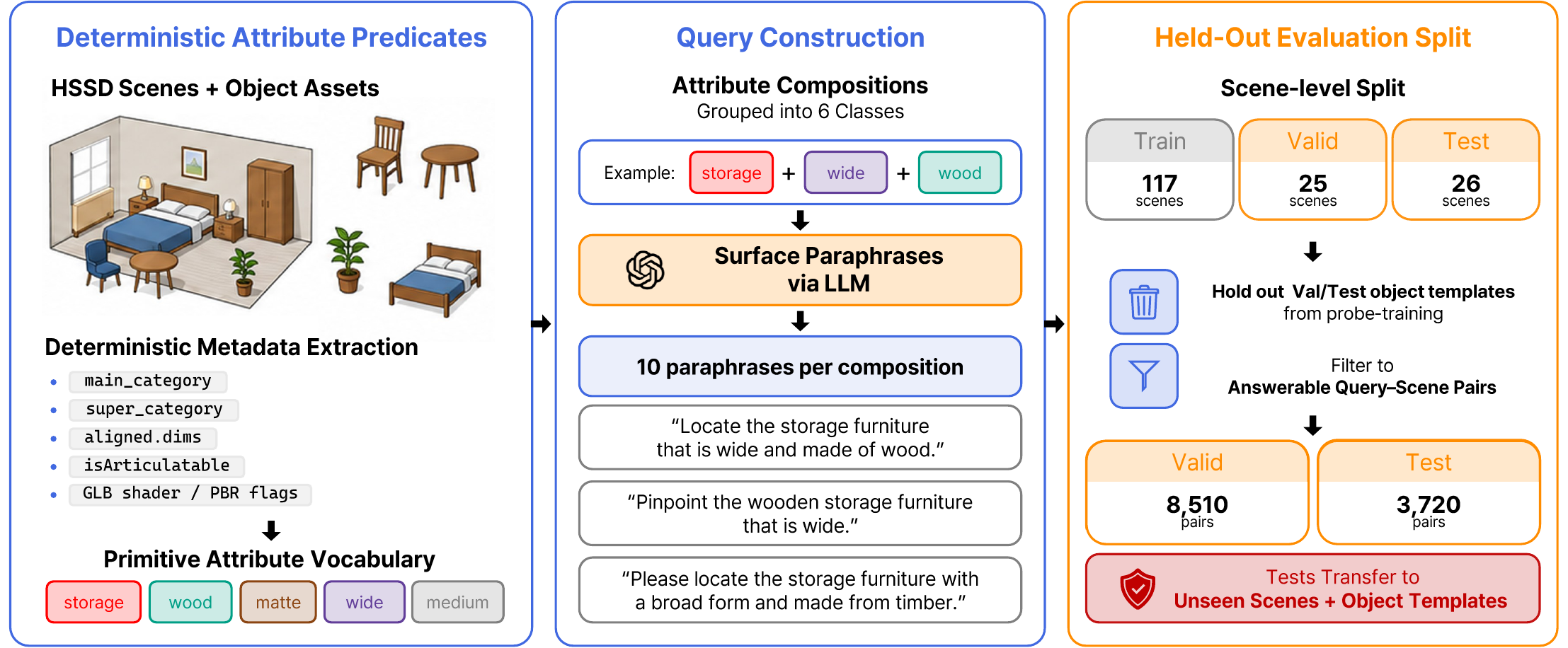}
\caption{\textbf{Attribute-Compositional Retrieval protocol construction.}
HSSD object annotations and asset metadata define a primitive vocabulary of category attributes and object properties, together with retrieval targets for each attribute composition.
GPT-4o generates natural-language paraphrases while preserving the underlying target specification.
Scene-, asset-, and composition-disjoint splits followed by answerability filtering produce the evaluation pairs.}
\label{fig:protocol}
\end{figure}

We instantiate the Attribute-Compositional Retrieval protocol on the Habitat Synthetic Scenes Dataset (HSSD)~\citep{khanna2024habitat} and construct it in three stages (Figure~\ref{fig:protocol}): (i) attribute formulation, which establishes a primitive attribute vocabulary and target predicates from object annotations and asset metadata, providing unambiguous retrieval targets; (ii) query construction, which converts attribute compositions into natural-language queries through constrained paraphrasing while preserving the underlying target specification; and (iii) evaluation split construction, which produces held-out evaluation pairs through scene- and object-asset-disjoint splits together with answerable-pair filtering, enabling evaluation on unseen scenes, object assets, and attribute compositions.

The resulting protocol isolates attribute-compositional retrieval: each query specifies a combination of object attributes, and a valid answer must satisfy the full composition.
All approaches use an identical object inventory, frozen object-node features, and scene-graph structure, removing differences in scene representation as a confound.

\paragraph{Attribute vocabulary.}
We define a primitive vocabulary $V$ of 47 attributes covering fine category, super category, material, size, surface roughness, metallicity, and articulatability.
The vocabulary is derived from HSSD annotations and asset metadata: fine categories and super categories come from \texttt{main\_category} and \texttt{super\_category}, size attributes from \texttt{aligned.dims}, articulatability from \texttt{isArticulatable}, and material, roughness, and metallicity from asset shader names and PBR flags.
We refer to material, size, surface roughness, metallicity, and articulatability as \emph{properties}; \emph{attribute} is used as an umbrella term encompassing both categories (fine and super categories) and properties.

\begin{table}[ht]
\centering
\small
\caption{Evaluation classes and retained query--scene pair counts. Validation contains seen compositions and test contains unseen compositions; difficulty statistics are averaged over answerable composition--scene combinations from both splits.}
\label{tab:eval_pairs}
\vspace{0.5em}
\resizebox{\textwidth}{!}{%
\begin{tabular}{lrrrrl}
\toprule
\multirow{2.5}{*}{\textbf{Class}}
& \multicolumn{2}{c}{\textbf{Query--scene pairs}}
& \multirow{2.5}{*}{\textbf{\shortstack{Targets\\/ scene}}}
& \multirow{2.5}{*}{\textbf{\shortstack{Near-miss\\/ scene}}}
& \multirow{2.5}{*}{\textbf{Example query}} \\
\cmidrule(lr){2-3}
& \textbf{Val. (seen)} & \textbf{Test (unseen)}
& & & \\
\midrule
C1: Fine cat.
& 1{,}190 & 750 & 4.60 & --
& \emph{``Find the chair.''} \\
C2: Fine cat. + material
& 1{,}030 & 160 & 2.03 & 18.65
& \emph{``Find a chair made of fabric.''} \\
C3: Fine cat. + size
& 2{,}260 & 190 & 2.52 & 33.29
& \emph{``Find the medium-sized ceiling lamp.''} \\
C4: Fine cat. + other property
& 530 & 200 & 2.60 & 49.53
& \emph{``Find the table with a glossy surface.''} \\
C5: Super cat. + one property
& 2{,}500 & 1{,}690 & 3.84 & 27.59
& \emph{``Find the storage furniture with a glossy finish.''} \\
C6: Three attributes
& 1{,}000 & 730 & 2.40 & 11.25
& \emph{``Find the large metal storage furniture.''} \\
\midrule
\textbf{Total} & \textbf{8{,}510} & \textbf{3{,}720} & & & \\
\bottomrule
\end{tabular}%
}
\end{table}

\paragraph{Composition classes and query construction.}
A target specification is a fixed conjunction of one to three attributes drawn from $V$; for example, \texttt{cat\_chair}$\land$\texttt{mat\_fabric} selects fabric chairs.
The benchmark defines 60 such attribute compositions, grouped into six classes (C1--C6) by the kind of conjunction required, as summarized in Table~\ref{tab:eval_pairs}.
For each query $q$, correct objects are those satisfying every attribute in the target specification.
The target specification is fixed before query generation; GPT-4o~\citep{hurst2024gpt} is used only offline to generate 10 natural-language paraphrases per composition under a structurally constrained prompt.
Thus, paraphrasing changes the surface form while leaving the target specification unchanged.
The full composition list is in Appendix~\ref{app:full-compositions}, and the paraphrasing prompt is in Appendix~\ref{app:prompt-paraphrase}.

\setlength{\columnsep}{8pt}
\begin{wraptable}[7]{r}{0.46\textwidth}
\vspace{-1.2em}
\centering
\scriptsize
\caption{Scene-level statistics for dataset splits.}
\label{tab:split_eval_stats}
\resizebox{\linewidth}{!}{%
\begin{tabular}{@{}lrrr@{}}
\toprule
 & \textbf{Train} & \textbf{Val} & \textbf{Test} \\
\midrule
Scenes                 & 117     & 25      & 26 \\
Objects / scene (avg.) & 109.9   & 105.2   & 118.5 \\
Rooms / scene (avg.)   & 5.32    & 5.20    & 5.15 \\
\bottomrule
\end{tabular}
}
\end{wraptable}

\paragraph{Scene-, asset-, and composition-disjoint split.}
We first split HSSD into 117 training, 25 validation, and 26 test scenes.
Because HSSD reuses 3D object assets across scenes, a scene-level split alone can place the same underlying asset in both training and evaluation scenes.
We therefore remove from the training pool every object asset that appears in a validation or test scene.
We additionally partition the 60 compositions into 43 seen compositions used for training and validation and 17 unseen compositions used only for testing.
Every primitive attribute in an unseen composition is instantiated by at least one object in both the training and validation scenes.
The test split therefore evaluates novel compositions of familiar attributes.
The unseen compositions are never used for validation or model selection.

\paragraph{Evaluation pairs and difficulty.}
We retain a query--scene pair only when the scene contains at least one object satisfying the target specification.
This yields 8{,}510 pairs for the 43 seen compositions across 25 validation scenes and 3{,}720 pairs for the 17 unseen compositions across 26 test scenes.
For each retained composition--scene combination, targets are objects satisfying the full specification, whereas near-misses are objects satisfying all but one required attribute.
Near-misses are therefore plausible distractors that can be rejected only by considering the full attribute conjunction.
Table~\ref{tab:eval_pairs} reports the average numbers of targets and near-misses within each class.


\section{FRAME}
\label{sec:frame}

\begin{figure}[h]
\centering
\includegraphics[width=0.98\linewidth,trim={0.5cm 0.0cm 0.5cm 0.5cm}]{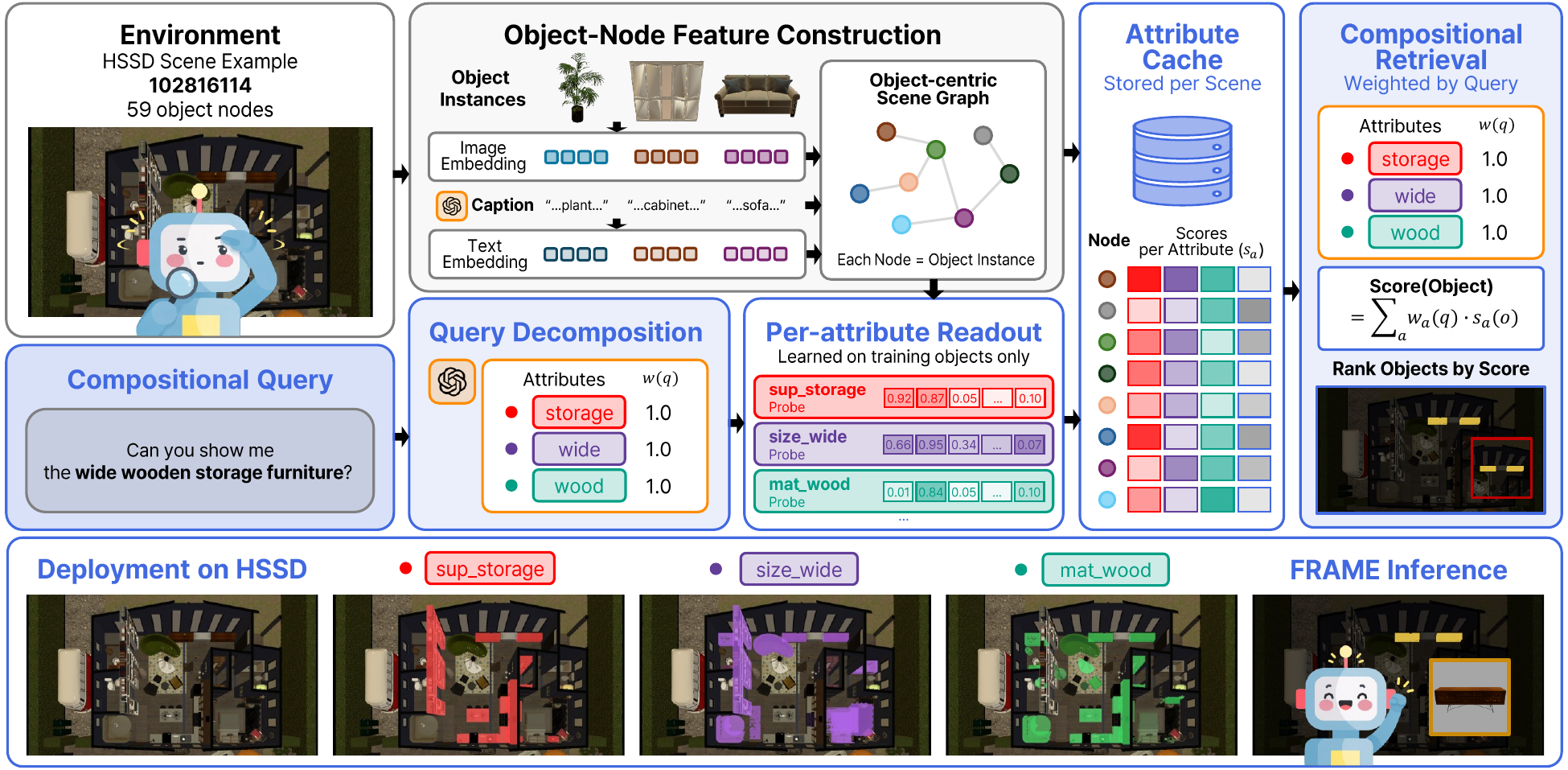}
\caption{\textbf{FRAME inference.}
A natural-language query is first decomposed into attribute weights over the primitive vocabulary.
For each object node in the scene memory, learned attribute readouts estimate evidence for the queried attributes from stored object embeddings.
The selected evidence is then aggregated according to the query weights to score and rank candidate objects.
The bottom row illustrates an HSSD test scene where \emph{storage furniture}, \emph{wide}, and \emph{wood} evidence combine to localize the target object.}
\label{fig:frame_overview}
\end{figure}

We propose FRAME as an approach to attribute-compositional retrieval under this protocol.
FRAME factors attribute-compositional retrieval into three parts:
(i) a language interface that specifies which attributes matter for a query, represented as a sparse composition vector over the primitive attribute vocabulary 
$V$,
(ii) trained attribute readouts that estimate per-object evidence for each attribute from object-node embeddings,
and (iii) a retrieval rule that composes the selected evidence according to the query.
This separation allows attribute evidence to be estimated once from stored object representations and reused across different queries through query-dependent composition.

\paragraph{Query understanding via LLM decomposition.}
At query time, a single LLM call converts a natural-language command $q$ into a sparse composition vector $w(q) \in [0,1]^{|V|}$ over the attribute vocabulary $V$. 
The decomposition prompt (Appendix~\ref{app:prompt-decompose}) instructs the LLM to assign higher weights to required attributes and lower weights to weak modifiers, omitting attributes that are irrelevant.
Each reference thereby defines a query-specific composition over persistent attributes; the resulting weights determine which object-side readouts are composed for retrieval.

\paragraph{Attribute probes.}
For each attribute $a \in V$, FRAME trains a 2-layer MLP readout probe $\phi_a$ on the attribute labels of objects $o$ in the training pool for a fixed number of epochs; architecture and training details are in Appendix~\ref{app:probe:arch}.
Given an object-node representation $\mathbf{x}(o)$, the probe outputs
\[
s_a(o) = \sigma(\phi_a(\mathbf{x}(o))) \in [0,1].
\]
We instantiate two variants: FRAME-Vis uses visual embeddings, $\mathbf{x}(o)=\evis(o)$, while FRAME-VT concatenates visual and caption-derived text features, $\mathbf{x}(o)=[\evis(o);\etext(o)]$.
In our approach, the validation split is used only for the readout-quality analysis in \S\ref{sec:experiments:probe-bottlenecks}, keeping the test split reserved for the retrieval evaluation in \S\ref{sec:experiments:main}.

\paragraph{Query-time retrieval.}
Given query-side composition weights $w(q)$ and per-attribute evidence scores, FRAME measures how well object $o$ satisfies the full attribute composition by a normalized weighted average:
\[
\mathrm{score}(o,q)
=
\frac{\sum_{a \in V} w_a(q)\,s_a(o)}
{\sum_{a \in V} w_a(q)},
\]
and returns the highest-scoring object node.
The probes can be applied \emph{lazily} at query time only for attributes with $w_a(q)>0$, or \emph{eagerly} at scene-ingestion time by caching all probe scores for a scene $s$ with objects $O_s$ as a per-scene matrix
\[
S_s \in [0,1]^{|O_s| \times |V|}.
\]
The eager mode trades a one-time ingestion cost for low-cost per-query object scoring; this is the setting used for the latency measurements in \S\ref{sec:experiments:main}.


\section{Experiments}
\label{sec:experiments}

\subsection{Experimental Setup}
\label{sec:experiments:setup}

\paragraph{Scene-memory representation.}
Evaluation uses 26 held-out test scenes containing 3{,}720 valid query--scene pairs for the 17 unseen compositions; validation uses 25 scenes containing 8{,}510 valid pairs for the 43 seen compositions.
All retrieval methods compared in this section operate on the same captioned object-centric scene graph, with one node per HSSD object in a ConceptGraph-style format~\citep{gu2024conceptgraphs}.
Each object node contains a \texttt{SigLIP-so400m} visual embedding~\citep{zhai2023sigmoid} ($\evis$, 1152d) from a canonical rendered view and a six-field GPT-4o-mini object caption (Appendix~\ref{app:prompt-caption}); caption-based methods encode captions with \texttt{BGE-large-en-v1.5}~\citep{bge_embedding} ($\etext$, 1024d).

\paragraph{Baselines.}
We compare representative approaches for retrieving objects from scene memory, spanning embedding-based retrieval, LLM-mediated ranking, retrieval-then-reranking pipelines, and attribute-compositional retrieval. 
These categories reflect different ways of accessing information stored in object-centric scene memories.

\begin{itemize}[leftmargin=1.2em, itemsep=1pt, topsep=2pt]

\item \textbf{Zero-shot embedding retrieval.}
Rank objects by direct query--object similarity without task-specific training.
ZS-Visual, ZS-Caption, and ZS-V+C use SigLIP text--image similarity, BGE query--caption similarity, and reciprocal-rank fusion of the two rankings, respectively.

\item \textbf{Learned embedding retrieval.}
Train two retrieval functions on the training pool with binary cross-entropy that scores positive pairs higher than negatives.
BiEncoder encodes the query and each object independently and ranks by a single cosine similarity, whereas CrossEncoder concatenates the query and object representations and predicts relevance jointly from each pair.

\item \textbf{LLM-mediated ranking.}
Use an LLM to rank objects directly from scene-memory contents, including captions and graph context. 
CG-LLM adapts ConceptGraphs~\citep{gu2024conceptgraphs}; BBQ~\citep{linok2025beyond} filters with short captions and reranks with full captions, 3D positions, and graph edges.

\item \textbf{Retrieval-then-LLM reranking.}
First retrieve a candidate set using embedding similarity, then use an LLM to rerank candidates with additional scene context.
EmbodiedRAG~\citep{booker2024embodiedrag} uses SigLIP top-$K$ retrieval with local-subgraph reranking; HOV-SG~\citep{werby2024hierarchical} additionally decomposes queries before filtering and reranking.

\item \textbf{Attribute-compositional retrieval.}
Decompose the query into attributes and retrieve objects by aggregating attribute-specific evidence rather than a single holistic similarity score.
FRAME-ZS replaces learned attribute probes with zero-shot visual--text similarity.
FRAME-Vis applies learned probes to SigLIP visual embeddings, while FRAME-VT applies them to concatenated SigLIP visual and BGE caption embeddings.
\end{itemize}

Unless otherwise noted, all LLM-based components, including FRAME query decomposition and LLM-mediated baselines, use \texttt{gpt-4o-mini} with temperature 0.
Prompts, top-$K$ values, model versions, and adaptation details are provided in Appendix~\ref{app:prompts}.

\paragraph{Evaluation metrics.}
For each query--scene pair, a method returns a ranked object list $R=(r_1,\ldots,r_{|O_s|})$ and is evaluated against the attribute-defined ground-truth set $G(q,s)$.
We use P@1 as the primary metric since the task requires returning a single object, and report Hit@5, mean reciprocal rank (MRR), and mean average precision (mAP) to characterize the full ranking:
\[
\mathrm{P@1}=\mathbb{1}[r_1\in G],\quad
\mathrm{Hit@5}=\mathbb{1}[\exists\,i\le 5:r_i\in G],
\]
\[
\mathrm{RR}=\frac{1}{\min\{i:r_i\in G\}},\quad
\mathrm{AP}=\frac{1}{|G|}\sum_{i=1}^{|O_s|}\mathrm{Prec@}i\cdot\mathbb{1}[r_i\in G],
\]
where $\mathrm{Prec@}i=\frac{1}{i}\sum_{j=1}^{i}\mathbb{1}[r_j\in G]$.
We average these quantities over all evaluated query--scene pairs and report them in percent.

\subsection{Compositional Retrieval Performance}
\label{sec:experiments:main}

\begin{table}[h]
\centering
\small
\setlength{\tabcolsep}{9.5pt}   
\caption{Compositional retrieval performance on the held-out test split, grouped by access pattern.
Accuracy metrics are reported in percent; latency is median per-query wall-clock time.
Best results are in \textbf{bold} and the best non-FRAME baseline is \underline{underlined}.}
\vspace{0.5em}
\label{tab:main}
\begin{tabular}{@{}llrrrrr@{}}
\toprule
Retrieval approach & Method & P@1 & Hit@5 & MRR & mAP & Latency \\
\midrule
\multirow{3}{*}{Zero-shot embedding retrieval}
  & ZS-Visual  & 46.40 & 83.92 & 62.22 & 51.57 & 10 ms \\
  & ZS-Caption & 30.16 & 72.58 & 48.33 & 37.79 & 13 ms \\
  & ZS-V+C     & 46.77 & 82.66 & 61.60 & 49.19 & 24 ms \\
\midrule
\multirow{2}{*}{Learned embedding retrieval}
  & BiEncoder    & 48.87 & 87.15 & 65.13 & 54.44 & 14 ms \\
  & CrossEncoder & \underline{62.58} & \underline{93.79} &
    \underline{75.50} & \underline{66.06} & 10 ms \\
\midrule
\multirow{2}{*}{LLM-mediated ranking}
  & CG-LLM & 27.23 & 66.53 & 43.96 & 29.66 & 1{,}996 ms \\
  & BBQ    & 41.21 & 75.13 & 55.17 & 38.11 & 6{,}256 ms \\
\midrule
\multirow{2}{*}{Retrieval-then-LLM reranking}
  & EmbodiedRAG & 44.57 & 78.68 & 59.10 & 48.20 & 3{,}187 ms \\
  & HOV-SG      & 42.02 & 77.88 & 57.50 & 45.64 & 4{,}398 ms \\
\midrule
\multirow{3}{*}{Attribute-compositional}
  & FRAME-ZS & 35.67 & 68.20 & 50.53 & 40.65 & 1{,}532 ms \\
  & \hc\textbf{FRAME-Vis}
    & \hc\textbf{77.82} & \hc\textbf{96.21}
    & \hc\textbf{86.01} & \hc\textbf{78.30} & \hc 1{,}638 ms \\
  & \hc\textbf{FRAME-VT}
    & \hc 75.75 & \hc 95.97 & \hc 84.62 & \hc 76.28 & \hc 1{,}750 ms \\
\bottomrule
\end{tabular}
\end{table}

\paragraph{Overall accuracy.}
Table~\ref{tab:main} shows that existing scene-memory retrieval methods fall short on attribute-compositional retrieval.
Zero-shot embedding retrieval is fast, but a single query--object similarity score is a weak interface for distinguishing full attribute matches from partial ones; the strongest variant, ZS-V+C, reaches 46.77\% P@1.
Even learned retrieval models (BiEncoder and CrossEncoder) reach only 48.87\% and 62.58\% P@1, respectively, suggesting that improved similarity or relevance learning alone does not resolve attribute-compositional retrieval.
LLM-mediated ranking and retrieval-then-LLM reranking remain below 44.57\% P@1 despite query-time reasoning over captions, graph context, or retrieved candidates.
FRAME-Vis achieves the best overall result at 77.82\% P@1, improving over CrossEncoder, the strongest non-FRAME baseline, by 15.24 percentage points.
FRAME-VT trails FRAME-Vis only slightly, at 75.75\% P@1.

\paragraph{Ranking quality beyond top-1.}
FRAME-Vis's advantage extends beyond P@1, also achieving the highest Hit@5, MRR, and mAP.
Notably, several baselines reach comparably high Hit@5 but leave a much larger gap to their own P@1; ZS-Visual reaches 83.92\% Hit@5 but only 46.40\% P@1, BiEncoder reaches 87.15\% Hit@5 bust only 48.87\% P@1, and CG-LLM reaches 66.53\% Hit@5 but only 27.23\% P@1.
This indicates that the correct object is often among the top few candidates across baselines, but resolving the full attribute conjunction to rank it first remains difficult without reliable, factored object-level evidence.

\paragraph{Factored retrieval and reliable readouts.}
These results suggest that factoring retrieval by attribute and estimating each attribute reliably interact rather than help independently.
FRAME-ZS shows that factoring alone does not help: it applies the same attribute-compositional interface as FRAME-Vis and FRAME-VT but uses zero-shot attribute scores instead of learned readouts, reaching only 35.67\% P@1, below even the holistic zero-shot baselines.
Conversely, BiEncoder and CrossEncoder show that task-tuning alone also has limits: both are trained on the same supervision as FRAME, yet each still collapses a multi-attribute reference into a single query--object score, and neither reaches 63\% P@1.
FRAME-Vis and FRAME-VT combine both ingredients, and the resulting gap over the best baseline is consistent with the two being complementary rather than separately sufficient.

\paragraph{Latency and retrieval cost.}
Embedding retrieval methods are fastest at 10--24~ms, but remain far behind FRAME-Vis and FRAME-VT in accuracy.
LLM-based ranking and reranking baselines require fresh LLM calls over scene or candidate context and take 2.0--6.3~s per query.
FRAME's total latency is also dominated by a single LLM call for query decomposition, making its absolute per-query latency comparable to the faster LLM-mediated baselines.
However, this call is scene-independent because its prompt contains the attribute vocabulary and the user query, not scene objects, captions, or candidate subgraphs.
Attribute readout scores are instead cached once per scene as an object--attribute matrix, so post-decomposition scoring reduces to a matrix--vector product taking approximately 0.01~ms.
Thus, FRAME's efficiency advantage is the decoupling of language parsing from scene-context reasoning and the amortization of attribute readout computation across repeated queries to the same memory.

\subsection{Performance Breakdown}
\label{sec:experiments:breakdown}

\paragraph{Probe bottlenecks.}
\label{sec:experiments:probe-bottlenecks}

\begin{wraptable}[8]{r}{0.350\textwidth}
\vspace{-1.2em}
\centering
\small
\caption{Probe validation AP over the 47 object attributes.}
\label{tab:probe_family_main}
\begin{tabular}{@{}lcc@{}}
\toprule
Group & AP\textsubscript{Vis} & AP\textsubscript{VT} \\
\midrule
Category        & 0.909 & 0.912 \\
Size            & 0.877 & 0.863 \\
Material        & 0.505 & 0.465 \\
Other properties & 0.840 & 0.821 \\
\bottomrule
\end{tabular}
\end{wraptable}

Because FRAME's retrieval score is built from per-attribute readouts, its accuracy should depend on how reliable those readouts are for the attributes a query requires.
We examine this by reporting validation AP for the learned readouts, grouped by attribute family (Table~\ref{tab:probe_family_main}).
Category and size readouts are reliable, readouts for other properties are moderately strong, and material readouts are substantially weaker.
Here, the \emph{category} group averages fine- and super-category probes, while \emph{other properties} comprise surface roughness, metallicity, and articulatability.
This suggests that material is the main bottleneck, likely because single-view material cues are confounded by texture, lighting, finish, and asset style.
This weakness is most directly reflected in class C2 (fine category + material), which explicitly requires material grounding.
Nevertheless, FRAME's gains on this class suggest that factored composition can combine weaker material evidence with other query-specified cues rather than collapsing the full reference into a single entangled score.

\begin{figure}[ht]
\centering
\includegraphics[width=\linewidth,trim={1.0cm 0.0cm 0.5cm 0.5cm}]{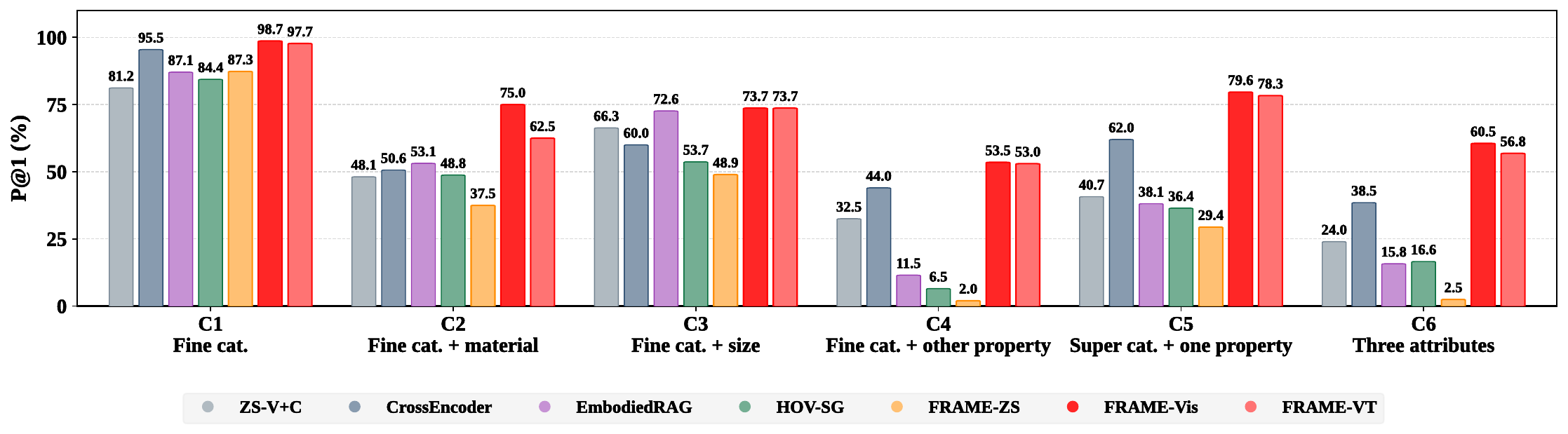}
\caption{\textbf{Per-class test P@1 (\%).} Representative baselines are shown for each retrieval family; the full table is in Appendix~\ref{app:full-results}. FRAME variants outperform the shown baselines across every composition class, with the largest gap on C6, where targets require conjunctions of three attributes.}
\label{fig:per_class}
\end{figure}

\paragraph{Per-class breakdown.}
Figure~\ref{fig:per_class} reports per-class P@1 for the representative baselines and FRAME variants.
The advantage of FRAME-Vis is smallest on single-category queries and generally larger on the multi-attribute classes.
CrossEncoder, the strongest non-FRAME baseline overall, reaches 95.5\% P@1 on single-category queries (C1) but falls to 38.5\% on three-attribute queries (C6), a 57.0,pp drop; the other baselines show the same broad decline from C1 to C6.
FRAME-Vis drops by 38.1\,pp over the same comparison ($98.7 \to 60.5$), showing robustness to conjunctive object references.

Comparing the three two-attribute classes reveals a consistent ordering across nearly all methods, including FRAME-ZS: size (C3) is easiest, material (C2) is harder, and the other-property family (C4) is hardest, with FRAME-ZS falling to 2.0\% on C4.
This ordering is not what the probe validation AP in Table~\ref{tab:probe_family_main} would predict on its own, since other-property probes are on average more reliable than material probes.
A likely factor is that C4 also has the highest near-miss density of the three classes (Table~\ref{tab:eval_pairs}), so ranking the correct object first requires distinguishing it from many more attribute-matching distractors, independent of how reliable any single attribute score is.
FRAME-Vis and FRAME-VT track each other closely across most classes but diverge on C2 (75.0\% vs. 62.5\%), consistent with material being the family where FRAME-VT's readouts are least reliable relative to FRAME-Vis (Table~\ref{tab:probe_family_main}).

\begin{figure}[ht]
\centering
\includegraphics[width=\linewidth,trim={1.0cm 0.5cm 0.5cm 0.5cm}]{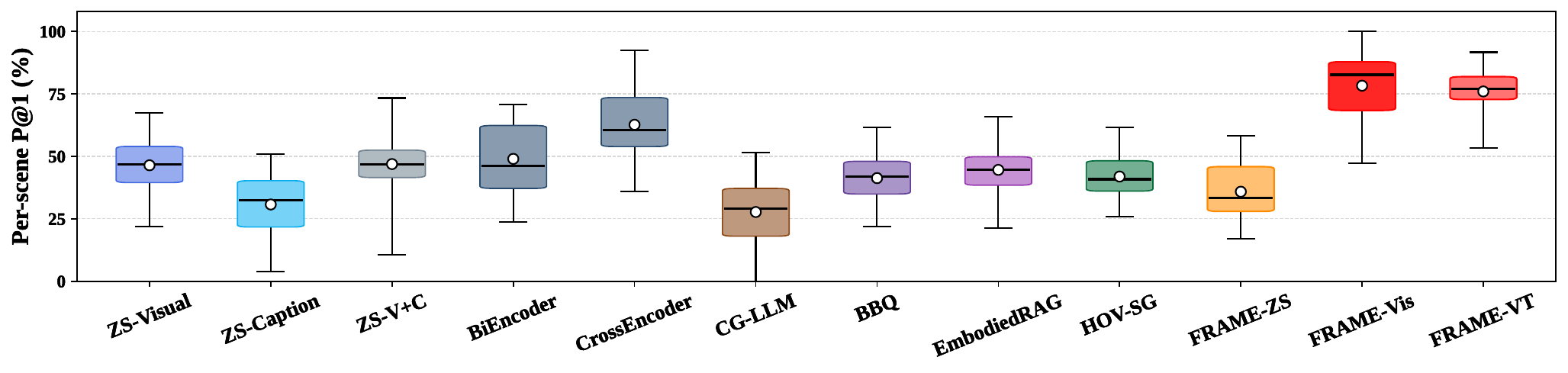}
\caption{\textbf{Per-scene test P@1 (\%) across all 26 test scenes.} Each box spans Q1--Q3 with the median marked by the horizontal line, whiskers extend to the minimum and maximum, and the white dot marks the mean. FRAME-Vis and FRAME-VT achieve the two highest medians, while FRAME-VT exhibits the tightest spread.}
\label{fig:per_scene_boxplot}
\end{figure}

\paragraph{Held-out scene consistency.}
To assess whether this held-out performance is consistent across individual scenes rather than driven by a few favorable ones, we examine per-scene P@1 across the 26 test scenes.
Figure~\ref{fig:per_scene_boxplot} reports the per-scene P@1 distribution for all methods.
FRAME-Vis maintains the highest median P@1 across scenes, indicating that its advantage holds consistently rather than concentrating in a few favorable layouts.
FRAME-VT trails FRAME-Vis slightly in overall P@1 but has the tightest spread across scenes (IQR 9.13, compared with 19.43 for FRAME-Vis and 19.61 for CrossEncoder).
This shows that FRAME's advantage holds consistently across scenes, rather than being driven by a small number of favorable layouts.

As a further, more demanding test of generalization, Appendix~\ref{app:real-scene-readout} reports results on the 3D Semantic Scene Graphs (3DSSG) dataset~\citep{wald2020learning}.
Rather than demonstrating deployment-ready real-scene grounding, these results show that semantic and appearance-related readouts can be learned from real-scene object image crops, while weak size readouts expose limitations arising from noisy geometry, the absence of depth or scale calibration, and imperfect segmentation.


\section{Conclusion}
\label{sec:conclusion}

This work positions attribute-compositional retrieval as a complementary mode of querying scene memories for language-guided robots.
FRAME embodies this view by letting language specify the persistent attributes that matter, reading out evidence for those attributes from object embeddings, and composing that evidence to retrieve objects matching multi-attribute references.
The controlled evaluation shows that this factorized retrieval strategy improves attribute-compositional retrieval over representative scene-memory retrieval baselines, while reducing post-decomposition object scoring to lightweight matrix-vector computation rather than repeated scene-context reasoning.
These results suggest that robot scene memories can benefit from compositional access to objects through language-specified persistent attributes, alongside existing spatial or relational querying.

\paragraph{Limitations.}
Our HSSD-based protocol isolates retrieval over a predefined attribute vocabulary, leaving open extensions to open-vocabulary attributes, relational references, and richer language phenomena.
Within this vocabulary, some attributes remain difficult to estimate reliably from a single view, most notably material, which limits how far factored composition alone can compensate for unreliable evidence.
Extending this perspective beyond the controlled setting and preliminary real-scene checks will require more robust robot-built object memories, including reliable segmentation, object-centric representation learning, and scene-graph construction.


\clearpage


\bibliography{references}  
\appendix
\newpage

\section{Probe Architecture and Training}
\label{app:probe-details}

\subsection{Architecture}
\label{app:probe:arch}

For each attribute $a \in V$, FRAME trains an independent two-hidden-layer MLP probe $\phi_a$ with hidden sizes $256$ and $128$, ReLU activations, and dropout $p=0.3/0.2$ after the two hidden layers.
The probe maps an object representation $\mathbf{x}(o)$ to a scalar logit, which is passed through a sigmoid at inference to produce

\[
s_a(o)=\sigma(\phi_a(\mathbf{x}(o)))\in[0,1].
\]

The input dimension is 1152 for FRAME-Vis and 2176 for FRAME-VT.
FRAME-VT concatenates separately $\ell_2$-normalized 1152-dimensional SigLIP visual and 1024-dimensional BGE caption embeddings.
The probes are independent across attributes and use neither parameter sharing nor auxiliary losses.

\subsection{Training procedure}
\label{app:probe:training}

For each attribute, we construct a binary classification problem over the asset-disjoint training pool.
Positive examples are training objects carrying the attribute label, and negatives are the remaining training objects.
We handle label imbalance using a positive-class weight in the binary cross-entropy loss:
\[
\mathcal{L}_a=-\left[w_+y\log\sigma(\phi_a(\mathbf{x}))+(1-y)\log\left(1-\sigma(\phi_a(\mathbf{x}))\right)\right],
\qquad w_+=N_{\mathrm{neg}}/N_{\mathrm{pos}}.
\]
Table~\ref{tab:probe-training-details} summarizes the fixed training configuration.

\begin{table}[!htbp]
\centering
\small
\caption{Probe training details.}
\label{tab:probe-training-details}
\begin{tabular*}{0.72\textwidth}{@{\extracolsep{\fill}}ll@{}}
\toprule
Item & Setting \\
\midrule
Training pool & Asset-disjoint training objects \\
Loss & BCE with logits \\
Class imbalance & Positive weight $w_+=N_{\mathrm{neg}}/N_{\mathrm{pos}}$ \\
Optimizer & Adam \\
Learning rate & $10^{-3}$ \\
Weight decay & $10^{-5}$ \\
Schedule & Cosine annealing, 80 epochs \\
Batch size & 512 \\
Probe inclusion & $\geq 30$ positive training objects \\
Hardware & NVIDIA RTX A6000 GPUs \\
\bottomrule
\end{tabular*}
\end{table}

All 47 object attributes used by the final protocol pass the inclusion threshold; the least frequent retained attribute has 66 positive training objects.

\subsection{Validation reporting}
\label{app:probe:selection}

Each probe is trained for a fixed 80 epochs, and the final-epoch checkpoint is used for retrieval evaluation.
We report per-attribute validation AP to characterize readout quality, but do not use validation AP for checkpoint selection.
Composition-level metrics such as P@1, Hit@5, MRR, and mAP are computed only when evaluating retrieval.

\subsection{Per-attribute validation AP}
\label{app:probe:per-attribute-ap}

Table~\ref{tab:per-attr-ap-full} reports validation AP for all 47 attribute probes under both input variants.
The grouped summary in Table~\ref{tab:probe_family_main} combines fine- and super-category probes as \emph{Category}, and roughness, metallicity, and articulatability as \emph{Other properties}.
Material is the weakest family, while the remaining attribute families are substantially stronger.
FRAME-VT has lower family-mean AP than FRAME-Vis for most attribute families, consistent with FRAME-Vis achieving the strongest overall retrieval performance.

\begin{table}[h]
\centering
\scriptsize
\caption{Per-attribute validation AP for FRAME-Vis and FRAME-VT, grouped by family. $\Delta=\mathrm{AP}_{\mathrm{VT}}-\mathrm{AP}_{\mathrm{Vis}}$.}
\label{tab:per-attr-ap-full}
\begin{tabular*}{\textwidth}{@{\extracolsep{\fill}}llrrrr@{}}
\toprule
Family & Attribute & AP$_{\mathrm{Vis}}$ & AP$_{\mathrm{VT}}$ & $\Delta$ & $N^+_{\mathrm{val}}$ \\
\midrule
\multicolumn{2}{l}{\textbf{Fine category} (19)} & \textbf{0.883} & \textbf{0.889} & \textbf{+0.006} & \\
 & \texttt{cat\_bed} & 1.000 & 1.000 & +0.000 & 54 \\
 & \texttt{cat\_couch} & 0.993 & 0.995 & +0.002 & 53 \\
 & \texttt{cat\_chair} & 0.991 & 0.993 & +0.001 & 102 \\
 & \texttt{cat\_potted\_plant} & 0.993 & 0.990 & -0.003 & 46 \\
 & \texttt{cat\_picture} & 0.979 & 0.977 & -0.002 & 172 \\
 & \texttt{cat\_ceiling\_lamp} & 0.979 & 0.971 & -0.007 & 77 \\
 & \texttt{cat\_carpet} & 0.949 & 0.953 & +0.004 & 122 \\
 & \texttt{cat\_curtain} & 0.946 & 0.947 & +0.001 & 39 \\
 & \texttt{cat\_table\_lamp} & 0.953 & 0.945 & -0.008 & 43 \\
 & \texttt{cat\_floor\_lamp} & 0.910 & 0.931 & +0.021 & 34 \\
 & \texttt{cat\_table} & 0.909 & 0.925 & +0.015 & 120 \\
 & \texttt{cat\_mirror} & 0.899 & 0.900 & +0.001 & 50 \\
 & \texttt{cat\_stool} & 0.822 & 0.895 & +0.073 & 13 \\
 & \texttt{cat\_chest\_of\_drawers} & 0.897 & 0.860 & -0.037 & 63 \\
 & \texttt{cat\_wardrobe} & 0.834 & 0.854 & +0.019 & 32 \\
 & \texttt{cat\_cabinet} & 0.790 & 0.805 & +0.015 & 79 \\
 & \texttt{cat\_shelves} & 0.736 & 0.748 & +0.012 & 55 \\
 & \texttt{cat\_wall\_lamp} & 0.733 & 0.700 & -0.033 & 16 \\
 & \texttt{cat\_counter} & 0.468 & 0.501 & +0.033 & 20 \\
\midrule
\multicolumn{2}{l}{\textbf{Super category} (11)} & \textbf{0.955} & \textbf{0.952} & \textbf{-0.003} & \\
 & \texttt{super\_sleeping\_furniture} & 1.000 & 1.000 & +0.000 & 54 \\
 & \texttt{super\_seating\_furniture} & 0.991 & 0.991 & -0.000 & 183 \\
 & \texttt{super\_plant} & 0.996 & 0.988 & -0.008 & 47 \\
 & \texttt{super\_decor} & 0.980 & 0.977 & -0.003 & 172 \\
 & \texttt{super\_lighting} & 0.980 & 0.977 & -0.003 & 171 \\
 & \texttt{super\_floor\_covering} & 0.951 & 0.956 & +0.005 & 122 \\
 & \texttt{super\_curtain} & 0.946 & 0.948 & +0.002 & 39 \\
 & \texttt{super\_bathroom\_fixtures} & 0.940 & 0.920 & -0.019 & 39 \\
 & \texttt{super\_storage\_furniture} & 0.913 & 0.912 & -0.001 & 249 \\
 & \texttt{super\_support\_furniture} & 0.902 & 0.901 & -0.001 & 135 \\
 & \texttt{super\_mirror} & 0.902 & 0.896 & -0.006 & 50 \\
\midrule
\multicolumn{2}{l}{\textbf{Material} (7)} & \textbf{0.505} & \textbf{0.465} & \textbf{-0.040} & \\
 & \texttt{mat\_metal} & 0.612 & 0.581 & -0.031 & 433 \\
 & \texttt{mat\_mirror} & 0.620 & 0.568 & -0.052 & 53 \\
 & \texttt{mat\_wood} & 0.583 & 0.546 & -0.037 & 262 \\
 & \texttt{mat\_rug} & 0.489 & 0.407 & -0.083 & 50 \\
 & \texttt{mat\_ceramic} & 0.493 & 0.394 & -0.099 & 30 \\
 & \texttt{mat\_leather} & 0.378 & 0.387 & +0.009 & 52 \\
 & \texttt{mat\_fabric} & 0.361 & 0.373 & +0.011 & 136 \\
\midrule
\multicolumn{2}{l}{\textbf{Size} (6)} & \textbf{0.877} & \textbf{0.863} & \textbf{-0.013} & \\
 & \texttt{size\_large} & 0.926 & 0.915 & -0.011 & 541 \\
 & \texttt{size\_low} & 0.904 & 0.893 & -0.011 & 177 \\
 & \texttt{size\_wide} & 0.898 & 0.884 & -0.014 & 437 \\
 & \texttt{size\_small} & 0.870 & 0.861 & -0.010 & 529 \\
 & \texttt{size\_tall} & 0.868 & 0.853 & -0.014 & 297 \\
 & \texttt{size\_medium} & 0.795 & 0.774 & -0.021 & 649 \\
\midrule
\multicolumn{2}{l}{\textbf{Roughness} (2)} & \textbf{0.734} & \textbf{0.703} & \textbf{-0.032} & \\
 & \texttt{roughness\_matte} & 0.816 & 0.784 & -0.032 & 577 \\
 & \texttt{roughness\_glossy} & 0.652 & 0.621 & -0.032 & 346 \\
\midrule
\multicolumn{2}{l}{\textbf{Metallicity} (1)} & \textbf{0.944} & \textbf{0.930} & \textbf{-0.013} & \\
 & \texttt{metallic} & 0.944 & 0.930 & -0.013 & 1177 \\
\midrule
\multicolumn{2}{l}{\textbf{Articulatability} (1)} & \textbf{0.949} & \textbf{0.950} & \textbf{+0.001} & \\
 & \texttt{articulatable} & 0.949 & 0.950 & +0.001 & 287 \\
\bottomrule
\end{tabular*}
\end{table}
\FloatBarrier

\section{Query Construction, Decomposition, and Baseline Details}
\label{app:prompts}

\subsection{FRAME query decomposition}
\label{app:prompt-decompose}

FRAME makes one scene-independent LLM call per query and supplies only the fixed 47-attribute vocabulary and the query text.
The output is projected onto the listed vocabulary before retrieval, so invalid keys and room-association attributes cannot enter the object score.

\textbf{Model.} \texttt{gpt-4o-mini}, \texttt{temperature=0}, \texttt{response\_format=\{"type":"json\_object"\}}.

\begin{promptbox}{FRAME query decomposition --- prompt}
You decompose object retrieval queries into attribute weights.

Vocabulary (47 attributes, grouped by type):

Object category:
  cat_chair, cat_table, cat_picture, cat_couch, cat_cabinet, cat_ceiling_lamp,
  cat_chest_of_drawers, cat_mirror, cat_bed, cat_potted_plant, cat_shelves,
  cat_table_lamp, cat_stool, cat_curtain, cat_counter, cat_wardrobe,
  cat_floor_lamp, cat_wall_lamp, cat_carpet

Super-category:
  super_storage_furniture, super_seating_furniture, super_decor, super_lighting,
  super_support_furniture, super_floor_covering, super_sleeping_furniture,
  super_plant, super_mirror, super_curtain, super_bathroom_fixtures

Size:
  size_small, size_medium, size_large, size_tall, size_low, size_wide

Material:
  mat_metal, mat_wood, mat_fabric, mat_mirror, mat_leather, mat_rug, mat_ceramic

Roughness/PBR:
  roughness_matte, roughness_glossy, metallic

Articulation:
  articulatable

For the query, output JSON with attribute weights in [0, 1]:
- 1.0 = required (must have)
- 0.7-0.9 = strongly preferred
- 0.3-0.5 = weakly preferred
- omit attributes that are irrelevant

Strict schema rules:
- Use only attribute names written verbatim in the vocabulary above.
- Use metallic, not roughness_metallic.
- Use super_bathroom_fixtures, not cat_bathroom_fixtures.
- Return only constraints stated by the query or unambiguously entailed by it.
- Do not add plausible alternatives, typical materials, sizes, or example categories.
- A generic request for seating furniture maps to super_seating_furniture, not to a list of subtypes.
- The request is conjunctive; do not turn one constraint into alternatives.
- Omit any attribute that is not confidently required.

Query: "{query}"

Output JSON: {"weights": {"attr_name": weight, ...}, "rationale": "brief"}
\end{promptbox}

\subsection{Paraphrase generation}
\label{app:prompt-paraphrase}

This procedure is run once at benchmark-construction time.
Each of the 60 active compositions is expanded into 10 natural-language requests, yielding 600 queries in total.
The eight C1 compositions use 10 direct requests each, whereas C2--C6 use seven direct and three lightly indirect requests per composition, producing 444 direct and 156 lightly indirect requests.
The generated requests are frozen and reused for all methods.

\textbf{Model.} \texttt{gpt-4o}, \texttt{temperature=0.65}, \texttt{response\_format=\{"type":"json\_object"\}}.

\begin{promptbox}{Paraphrase generation --- prompt}
Write natural, standalone English requests for a home robot to locate an
object in its current scene.

The exact target is: {description}
Every request MUST contain this exact head noun: "{head_noun}".

Return exactly {n_direct} direct requests and {n_indirect} light-indirect requests.
Direct requests state the attributes plainly. Light-indirect requests may paraphrase
one or more NON-NOUN attributes using ordinary physical language (for example,
"wooden" -> "crafted from timber", "glossy" -> "with a reflective sheen", and
"articulatable" -> "with movable parts"). They must still require exactly every stated
property. Do not turn the object class into a broader class, a narrower subtype, or a
functional description.

Never add an unguaranteed fact: no colour, room/location, nearby object, owner, use
history, object contents, construction details (doors, drawers, handles, panels),
comparatives, or superlatives. Never say "scene", "room", "view", "here", "nearby",
"hanging", "compartment", or "wall" unless it is in the exact head noun. {metal_note}
Do not mention labels, attributes,
annotations, or a benchmark. No riddles.

Return JSON only: {"direct": ["..."], "indirect": ["..."]}
\end{promptbox}

\noindent \texttt{\{description\}} is a deterministic textual rendering of the composition, and \texttt{\{head\_noun\}} is its category or super-category noun.
For C1, \texttt{n\_direct}=10 and \texttt{n\_indirect}=0; for C2--C6, they are 7 and 3, respectively.
When a composition requires \texttt{mat\_metal} but not \texttt{metallic}, \texttt{\{metal\_note\}} instructs the model not to conflate the material with the metallic finish.
Generated requests are accepted only if they contain the fixed head noun and pass the lexical safety checks; failed generations are retried up to three times.

\subsection{Per-object caption}
\label{app:prompt-caption}

Each object asset is captioned offline from its isolated object rendering.
The resulting caption is stored once per asset and reused as shared object-text memory by all caption-based methods.

\textbf{Model.} \texttt{gpt-4o-mini} with image input, \texttt{temperature=0.3}, \texttt{detail=low}, and a maximum of 400 output tokens.

\begin{promptbox}{Per-object caption --- prompt}
You are writing for a high-end furniture catalog. Describe the object in
this image using exactly these fields, in this format:

Type: [object category and sub-type]
Style: [aesthetic style, era, or cultural origin]
Materials: [primary and secondary materials, including texture and finish]
Features: [distinguishing parts and details you can see]
Atmosphere: [mood the object creates and ideal setting]
Descriptors: [3-5 evocative adjectives capturing the feel]

Rules:
- 50-80 characters per field, full descriptive phrases (not just keywords)
- Each object must have UNIQUE descriptions specific to what you actually see
- Describe ONLY the object, not the background or image quality
- Do NOT mention rendering, 3D, digital aspects, photographs, or backgrounds
- Use rich, evocative language as if writing for a luxury furniture catalog
\end{promptbox}

\noindent The same full caption is embedded for ZS-Caption, ZS-V+C, and the FRAME-VT object representation.
For the LLM-mediated baselines, \texttt{short\_caption(...)} extracts the \texttt{Type:} field: CG-LLM and BBQ Stage~1 use this short form, while BBQ Stage~2, EmbodiedRAG candidate reranking, and HOV-SG use the full caption.
EmbodiedRAG additionally uses the short form for neighboring objects in each candidate's local subgraph.

\subsection{BiEncoder}
\label{app:biencoder}

BiEncoder is a task-tuned embedding-retrieval baseline using the same frozen visual object features, asset-disjoint training pool, 43 seen compositions, frozen query forms, and evaluation harness as the other methods in our protocol.

\textbf{Architecture.} The query encoder is frozen BGE-large (\texttt{bge-large-en-v1.5}, CLS token, $\ell_2$-normalized), followed by a learnable projection $g_q$.
The object encoder applies a learnable projection $g_o$ to the frozen 1152-dimensional visual representation $\mathbf e_{\mathrm{vis}}(o)$.
Both projections are three-layer MLPs with dimensions $d_{\mathrm{in}}\!\to\!512\!\to\!256\!\to\!256$, ReLU activations, and dropout 0.3 after each hidden layer.
The projected vectors are $\ell_2$-normalized, and the retrieval score is
\[
\mathrm{score}(o,q)=\cos\!\left(g_o(\mathbf e_{\mathrm{vis}}(o)),g_q(\mathrm{BGE}(q))\right).
\]

\textbf{Training.} For each seen-composition--scene unit from the asset-disjoint training pool, positives are all objects satisfying every required attribute.
Negatives include at most four near-miss objects that satisfy all but one required attribute and at most four randomly sampled non-positive objects.
One frozen query form is sampled for each unit in every epoch.
BiEncoder is trained for a fixed 60 epochs using binary cross-entropy on the scaled cosine logit $10\,\mathrm{score}(o,q)$, Adam with learning rate $10^{-3}$ and weight decay $5\times10^{-4}$, cosine learning-rate scheduling, and batch size 1024.
The final-epoch checkpoint is used for evaluation; validation P@1 is not used for checkpoint selection.

\textbf{Inference.} Projected object vectors are precomputed and cached.
Each query requires one BGE encoding, one query projection, and one cosine matrix--vector product against the cached object matrix.

\subsection{CrossEncoder}
\label{app:crossencoder}

CrossEncoder provides a task-tuned relevance baseline under the same object features, training pool, query forms, supervision, and fixed 60-epoch schedule as BiEncoder.

\textbf{Architecture.} CrossEncoder concatenates the frozen 1024-dimensional BGE query representation with the frozen 1152-dimensional visual object representation.
A three-layer MLP with dimensions $2176\!\to\!512\!\to\!256\!\to\!1$, ReLU activations, and dropout 0.3 after each hidden layer maps the concatenation to a scalar relevance logit.

\textbf{Training.} CrossEncoder uses the same positive, near-miss, and random-negative sampling as BiEncoder and is optimized directly with binary cross-entropy on its relevance logit.
The optimizer, learning-rate schedule, batch size, and final-epoch checkpoint policy are identical to those of BiEncoder.

\textbf{Inference.} Each query representation is concatenated with every candidate object representation and scored by the MLP.
Consequently, unlike BiEncoder, CrossEncoder produces a query-dependent object score and does not admit a standalone cached object retrieval vector.

\subsection{FRAME-ZS attribute prompts}
\label{app:prompt-framezs}

FRAME-ZS uses the same strict decomposition and weighted composition interface as FRAME-Vis, but replaces learned probes with SigLIP text--image cosine scores.
The performance evaluation uses deterministic type-dependent templates rather than a hand-written per-attribute prompt dictionary.
Fine categories use \texttt{a photo of a/an <category>}; coarse categories use the corresponding singular object noun; materials use \texttt{a photo of an object made of <material>}; and size and physical-property attributes use \texttt{a photo of a/an <attribute> object}.
No room-association prompt is used.
For decomposed weights $\{w_a\}$, FRAME-ZS scores object $o$ by
\[
\mathrm{score}(o,q)=\frac{1}{\sum_a w_a}\sum_a w_a\,\cos\!\left(\mathbf e_{\mathrm{vis}}(o),\mathrm{SigLIP}_{\mathrm{text}}(\tau(a))\right).
\]

\subsection{CG-LLM}
\label{app:prompt-cgllm}

\textbf{Original framework.} ConceptGraphs~\citep{gu2024conceptgraphs} constructs an open-vocabulary 3D scene graph from RGB-D observations through object detection, tracking, and inter-object relation extraction, and uses LLM-based planning over the resulting scene graph for tasks such as task planning, question answering, and embodied action.

\textbf{Our adaptation.} We isolate the LLM grounding component: the scene graph and object features are provided by our shared protocol (\S\ref{sec:protocol}), and a single LLM call ranks objects from each object's short caption and up to three scene-graph neighbors. The full task-planning and structured-reasoning pipeline of the original is replaced with a ranked-list output. \texttt{gpt-4o-mini} replaces GPT-4.

\textbf{Model.} \texttt{gpt-4o-mini}, \texttt{temperature=0}, max 200 output tokens. $\textsc{TOP\_K} = 20$.

\begin{promptbox}{CG-LLM --- prompt}
You are an object retrieval assistant working with a 3D scene graph.
Scene has {N} objects, each with a short type and scene-graph neighbors:

1. {short_caption_1}  [neighbors: {rel_1} a {nbr_type_1}; ...]
2. {short_caption_2}  [neighbors: ...]
...
N. {short_caption_N}  [neighbors: ...]

User wants: "{query}"

Return the top {TOP_K} object indices most likely to match, ranked best->worst.
Output ONLY comma-separated 1-based indices.
\end{promptbox}

\subsection{BBQ}
\label{app:prompt-bbq}

\textbf{Original framework.} BBQ (Beyond Bare Queries)~\citep{linok2025beyond} constructs its own object-centric 3D scene graph from RGB-D scans, then performs grounding via a coarse caption-based filter followed by a fine reranking step with full captions, 3D positions, and graph edges. The framework additionally supports spatial-anchor queries (``X near Y'') through an anchor-resolution module.

\textbf{Our adaptation.} The scene graph and object features are provided by our shared protocol; the two-stage filter-and-rerank pipeline is preserved. The anchor-resolution logic is present in the released code but not exercised here, since our queries are attribute-compositional rather than spatial-anchor.

\textbf{Model.} \texttt{gpt-4o-mini}, \texttt{temperature=0}. $\textsc{STAGE1\_MAX\_KEEP} = 15$. Stage 1 max 150 tokens; Stage 2 max 300.

\begin{promptbox}{BBQ --- Stage 1 prompt}
Scene with {N} objects (short types):

1. {short_caption_1}
2. {short_caption_2}
...
N. {short_caption_N}

User wants: "{query}"

Select up to {STAGE1_MAX_KEEP} candidate indices likely to match.
Output ONLY comma-separated 1-based indices.
\end{promptbox}

\begin{promptbox}{BBQ --- Stage 2 prompt}
Query: "{query}"

Candidates with full captions, 3D positions, and scene-graph edges
(with reasoning):

1. {full_caption_1}
   Position: ({cx},{cy},{cz})/extent({ex},{ey},{ez})
   Edges: {rel_1} a {tag_1} --- {reasoning_1[:60]}; ...
2. ...
...
K. ...

Rank ALL {K} candidates by relevance, best first.
Output ONLY comma-separated 1-based indices (include all).
\end{promptbox}

\subsection{EmbodiedRAG}
\label{app:prompt-erag}

\textbf{Original framework.} EmbodiedRAG~\citep{booker2024embodiedrag} dynamically retrieves query-relevant 3D scene \emph{subgraphs} from a continuously updated scene graph to augment an LLM-based planner for robot task execution. The retrieval reduces input token counts so the planner can operate at scale; the evaluation target is task success in simulated household environments.

\textbf{Our adaptation.} The scene graph is provided by our shared (static) protocol. The retrieval unit is reduced from \emph{subgraph} to \emph{object node}, and the LLM step is changed from planner to object ranker. Top-$K$ object candidates are first retrieved by SigLIP visual--text cosine, then the LLM reranks them using each candidate's full caption and a 1-hop subgraph---preserving the local-subgraph reasoning that motivates the original retrieval.

\textbf{Model.} \texttt{gpt-4o-mini}, \texttt{temperature=0}, max 200 tokens. $\textsc{TOPK} = 20$, $\textsc{NEIGHBORS} = 3$ per candidate.

\begin{promptbox}{EmbodiedRAG --- prompt}
Query: "{query}"

Top {K} candidates with their local subgraphs (scene-graph neighbors):

1. {full_caption_1}
   Local subgraph: {rel_1} {nbr_type_1}; {rel_2} {nbr_type_2}; ...
2. ...
...
K. ...

Rank ALL {K} candidates by relevance, best first.
Output ONLY comma-separated 1-based indices.
\end{promptbox}

\subsection{HOV-SG}
\label{app:prompt-hovsg}

\textbf{Original framework.} HOV-SG~\citep{werby2024hierarchical} builds a hierarchical open-vocabulary 3D scene graph organized as floor $\to$ region $\to$ object, and uses this multi-level structure for language-grounded robot navigation by decomposing queries into floor / room / category / attribute components and traversing the hierarchy.

\textbf{Our adaptation.} The scene graph and object features are provided by our shared protocol. Since HSSD scenes are single-floor residential environments, the original floor--region--object hierarchy is reduced to room--object. The two-step pipeline (LLM decomposition into \texttt{\{room, category, attributes\}}, followed by reranking on room-and-category-filtered candidates) is preserved. The navigation-execution component of the original is not used; we only evaluate object-grounding ranking.

\textbf{Model.} \texttt{gpt-4o-mini}, \texttt{temperature=0}. Decompose max 100 tokens; rerank max 250. $\textsc{TOPK\_CAT} = 30$.

\begin{promptbox}{HOV-SG --- Step 1 decomposition prompt}
Decompose this object query into structured fields.
Query: "{query}"

Output JSON with three fields:
- room: the most likely room (bedroom, bathroom, kitchen, living_room,
  dining_room, office, kids_room, or "any" if unspecified)
- category: the main object class (chair, lamp, table, ...)
- attributes: list of descriptive words (e.g., ["wooden", "small"])

Output ONLY valid JSON. Example:
{"room": "bedroom", "category": "bed", "attributes": ["large"]}
\end{promptbox}

\begin{promptbox}{HOV-SG --- Step 2 reranking prompt}
Original query: "{query}"
Hierarchical decomposition: room={room}, category={category},
                            attributes={attrs_csv}

Candidates (already filtered by room + category):
1. {full_caption_1}
2. {full_caption_2}
...
K. {full_caption_K}

Rank ALL {K} candidates by relevance to the original query, best first.
Output ONLY comma-separated 1-based indices.
\end{promptbox}

\section{Active Attribute Compositions}
\label{app:full-compositions}

Table~\ref{tab:full-compositions} lists the 60 active compositions in the final protocol.
A dagger denotes one of the 17 unseen test compositions; the remaining 43 compositions provide training supervision and are evaluated on the validation scenes.
Every composition has 10 frozen query forms.

\begin{table}[!htbp]
\centering
\scriptsize
\caption{The 60 active attribute compositions. $\dagger$ denotes an unseen test composition.}
\label{tab:full-compositions}
\begin{tabularx}{\textwidth}{@{}lX@{}}
\toprule
Class & Active compositions \\
\midrule
C1: Fine cat. ($n=8$) &
\texttt{cabinet}, \texttt{carpet}, \texttt{ceiling\_lamp}$^\dagger$, \texttt{chair}$^\dagger$, \texttt{couch}, \texttt{mirror}, \texttt{potted\_plant}, \texttt{table}$^\dagger$ \\
\addlinespace[2pt]
C2: Fine cat. + material ($n=7$) &
\texttt{fabric\_chair}$^\dagger$, \texttt{leather\_chair}, \texttt{metal\_ceiling\_lamp}, \texttt{metal\_chair}, \texttt{metal\_table}, \texttt{wood\_shelves}, \texttt{wood\_table} \\
\addlinespace[2pt]
C3: Fine cat. + size ($n=12$) &
\texttt{large\_cabinet}, \texttt{large\_chair}, \texttt{large\_potted\_plant}, \texttt{large\_table}, \texttt{medium\_carpet}, \texttt{medium\_ceiling\_lamp}$^\dagger$, \texttt{medium\_picture}, \texttt{medium\_potted\_plant}, \texttt{medium\_table}, \texttt{tall\_cabinet}, \texttt{tall\_potted\_plant}, \texttt{wide\_table} \\
\addlinespace[2pt]
C4: Fine cat. + other property ($n=4$) &
\texttt{glossy\_table}$^\dagger$, \texttt{matte\_table}, \texttt{metallic\_cabinet}, \texttt{metallic\_ceiling\_lamp} \\
\addlinespace[2pt]
C5: Super cat. + one property ($n=19$) &
\texttt{ceramic\_bathroom\_fixtures}, \texttt{fabric\_seating\_furniture}, \texttt{glossy\_storage\_furniture}$^\dagger$, \texttt{large\_plant}, \texttt{large\_seating\_furniture}, \texttt{large\_storage\_furniture}$^\dagger$, \texttt{matte\_storage\_furniture}, \texttt{medium\_plant}, \texttt{medium\_seating\_furniture}, \texttt{medium\_storage\_furniture}$^\dagger$, \texttt{metal\_bathroom\_fixtures}, \texttt{metal\_seating\_furniture}, \texttt{metal\_storage\_furniture}$^\dagger$, \texttt{tall\_plant}, \texttt{tall\_storage\_furniture}$^\dagger$, \texttt{wide\_seating\_furniture}, \texttt{wide\_storage\_furniture}$^\dagger$, \texttt{wood\_seating\_furniture}, \texttt{wood\_storage\_furniture}$^\dagger$ \\
\addlinespace[2pt]
C6: Three attributes ($n=10$) &
\texttt{large\_glossy\_storage\_furniture}, \texttt{large\_metal\_storage\_furniture}$^\dagger$, \texttt{medium\_fabric\_seating\_furniture}, \texttt{medium\_glossy\_storage\_furniture}$^\dagger$, \texttt{medium\_metal\_storage\_furniture}, \texttt{medium\_wood\_storage\_furniture}, \texttt{tall\_glossy\_storage\_furniture}, \texttt{tall\_metal\_storage\_furniture}$^\dagger$, \texttt{wide\_metal\_storage\_furniture}, \texttt{wide\_wood\_storage\_furniture}$^\dagger$ \\
\bottomrule
\end{tabularx}
\end{table}

\section{Full Results}
\label{app:full-results}

\subsection{Full test metric breakdown}
\label{app:full-metrics}

Table~\ref{tab:full-main} reports all ranking metrics on the 17 unseen compositions and 26 held-out test scenes.
We do not mix these results with validation performance: validation scenes contain the 43 seen compositions and are used only for development diagnostics and the readout-quality analysis above.

\begin{table}[!htbp]
\centering
\small
\caption{Retrieval performance on 3{,}720 valid query--scene pairs from the unseen-composition test split. Values are percentages. Best results are in \textbf{bold}, and the best non-FRAME result is \underline{underlined}.}
\label{tab:full-main}
\begin{tabular*}{\textwidth}{@{\extracolsep{\fill}}lrrrr@{}}
\toprule
Method & P@1 & Hit@5 & MRR & mAP \\
\midrule
ZS-Visual & 46.40 & 83.92 & 62.22 & 51.57 \\
ZS-Caption & 30.16 & 72.58 & 48.33 & 37.79 \\
ZS-V+C & 46.77 & 82.66 & 61.60 & 49.19 \\
\midrule
BiEncoder & 48.87 & 87.15 & 65.13 & 54.44 \\
CrossEncoder & \underline{62.58} & \underline{93.79} & \underline{75.50} & \underline{66.06} \\
\midrule
CG-LLM & 27.23 & 66.53 & 43.96 & 29.66 \\
BBQ & 41.21 & 75.13 & 55.17 & 38.11 \\
EmbodiedRAG & 44.57 & 78.68 & 59.10 & 48.20 \\
HOV-SG & 42.02 & 77.88 & 57.50 & 45.64 \\
\midrule
FRAME-ZS & 35.67 & 68.20 & 50.53 & 40.65 \\
\textbf{FRAME-Vis} & \textbf{77.82} & \textbf{96.21} & \textbf{86.01} & \textbf{78.30} \\
FRAME-VT & 75.75 & 95.97 & 84.62 & 76.28 \\
\bottomrule
\end{tabular*}
\end{table}

\subsection{Per-class test P@1}
\label{app:full-per-class}

Table~\ref{tab:full-per-class} reports P@1 for all methods under the six composition classes.
FRAME-Vis achieves the highest P@1 in five of the six classes; BBQ is highest on C3.

\begin{table}[!htbp]
\centering
\scriptsize
\caption{Per-class test P@1 (\%) on the unseen-composition split. Best results are in \textbf{bold}, and the best non-FRAME result in each class is \underline{underlined}.}
\label{tab:full-per-class}
\begin{tabular*}{\textwidth}{@{\extracolsep{\fill}}lrrrrrr@{}}
\toprule
Method & C1 & C2 & C3 & C4 & C5 & C6 \\
\midrule
ZS-Visual & 78.53 & \underline{59.38} & 65.26 & 38.00 & 39.35 & 24.25 \\
ZS-Caption & 48.27 & 25.62 & 23.68 & 15.00 & 31.30 & 15.75 \\
ZS-V+C & 81.20 & 48.12 & 66.32 & 32.50 & 40.71 & 23.97 \\
\midrule
BiEncoder & 88.53 & 35.62 & 68.42 & 13.00 & 42.84 & 29.73 \\
CrossEncoder & \underline{95.47} & 50.62 & 60.00 & \underline{44.00} & \underline{62.01} & \underline{38.49} \\
\midrule
CG-LLM & 37.73 & 30.00 & 43.16 & 6.00 & 27.93 & 15.89 \\
BBQ & 68.93 & 41.25 & \textbf{\underline{77.89}} & 4.50 & 38.82 & 18.77 \\
EmbodiedRAG & 87.07 & 53.12 & 72.63 & 11.50 & 38.11 & 15.75 \\
HOV-SG & 84.40 & 48.75 & 53.68 & 6.50 & 36.45 & 16.58 \\
\midrule
FRAME-ZS & 87.33 & 37.50 & 48.95 & 2.00 & 29.41 & 2.47 \\
\textbf{FRAME-Vis} & \textbf{98.67} & \textbf{75.00} & 73.68 & \textbf{53.50} & \textbf{79.64} & \textbf{60.55} \\
FRAME-VT & 97.73 & 62.50 & 73.68 & 53.00 & 78.34 & 56.85 \\
\bottomrule
\end{tabular*}
\end{table}
\FloatBarrier

\section{Real-scene Attribute Readouts on 3DSSG}
\label{app:real-scene-readout}

Our main protocol fixes perception and scene-graph construction in order to isolate the access layer of object-centric grounding: given a fixed scene memory, how well can attribute-compositional language retrieve object nodes?
As a small real-scene sanity check, we evaluate learned attribute readouts on 3DSSG~\citep{wald2020learning}, using object crops extracted from 20 real-scene scans.
This experiment is not intended as a full end-to-end real-scene grounding
benchmark; rather, it tests whether the same attribute-readout architecture
can extract useful signal from noisier real-scene object crops.

\paragraph{Setup.}
We select 20 3DSSG scenes with relatively rich attribute coverage and use a fixed stratified split of 12/4/4 scenes for train/validation/test.
The resulting split contains 1,204 object instances in total, with 739/217/248 objects in train/validation/test.
We map 3DSSG object and attribute annotations into the 47-object-attribute vocabulary used by the final HSSD protocol; 21 attributes have at least 10 positive training examples and are retained for probe training.
For each oriented bounding-box instance, we extract an RGB crop by selecting the best available view according to visibility and Laplacian sharpness over the captured frames.
We encode each crop with the same SigLIP-so400m visual encoder used in the main experiments and train one MLP probe per attribute using the same probe architecture as Appendix~\ref{app:probe-details}.
We report mean and standard deviation over five training seeds; the split is fixed, so the seed varies probe initialization and minibatch order.

\paragraph{Family-level results.}
Table~\ref{tab:3dssg-family} summarizes test mean AP by attribute family.
All families are above their random baselines, but the performance pattern is uneven.
Super-category, material, and category readouts retain meaningful real-domain signal, while size readouts are close to random.

\begin{table}[h]
\centering
\small
\caption{3DSSG crop-level sanity check: family-level test mean AP over retained probes. Mean and standard deviation are computed over five training seeds.}
\label{tab:3dssg-family}
\begin{tabular}{lrrrr}
\toprule
Family & mAP & Std. & \# probes & Random AP \\
\midrule
Super-category & 0.555 & 0.024 & 6 & 0.062 \\
Material       & 0.494 & 0.052 & 3 & 0.078 \\
Category       & 0.456 & 0.009 & 7 & 0.046 \\
Size           & 0.132 & 0.005 & 5 & 0.060 \\
\bottomrule
\end{tabular}
\end{table}

\paragraph{Per-attribute pattern.}
The result is mixed but informative.
Several readouts transfer well to real-scene crops, especially attributes with distinctive visual signatures: \texttt{super\_plant} reaches 0.884 AP, \texttt{cat\_potted\_plant} 0.893, \texttt{super\_lighting} 0.770, \texttt{mat\_fabric} 0.684, and \texttt{cat\_table} 0.673.
Other readouts degrade substantially, including \texttt{size\_small} 0.054, \texttt{size\_large} 0.083, \texttt{size\_wide} 0.122, \texttt{cat\_couch} 0.202, and \texttt{cat\_cabinet} 0.207.
Low-positive-count attributes also show higher variance, for example \texttt{mat\_metal} and \texttt{super\_decor}, indicating that part of the instability comes from the small real-scene test pool.

\begin{table}[h]
\centering
\small
\caption{Selected 3DSSG per-probe test AP values, showing representative robust and weak readouts.}
\label{tab:3dssg-selected}
\begin{tabular}{lrrrr}
\toprule
Probe & AP & Std. & Train $N^+$ & Test $N^+$ \\
\midrule
\multicolumn{5}{l}{\textbf{Robust readouts}} \\
\texttt{cat\_potted\_plant} & 0.893 & 0.030 & 44 & 7 \\
\texttt{super\_plant} & 0.884 & 0.003 & 44 & 7 \\
\texttt{super\_lighting} & 0.770 & 0.032 & 18 & 7 \\
\texttt{mat\_fabric} & 0.684 & 0.032 & 104 & 35 \\
\texttt{cat\_table} & 0.673 & 0.026 & 44 & 17 \\
\midrule
\multicolumn{5}{l}{\textbf{Weaker readouts}} \\
\texttt{cat\_cabinet} & 0.207 & 0.022 & 15 & 6 \\
\texttt{cat\_couch} & 0.202 & 0.025 & 19 & 9 \\
\texttt{size\_wide} & 0.122 & 0.020 & 45 & 12 \\
\texttt{size\_large} & 0.083 & 0.011 & 19 & 5 \\
\texttt{size\_small} & 0.054 & 0.005 & 17 & 5 \\
\bottomrule
\end{tabular}
\end{table}

\paragraph{Two layers of failure.}
We interpret the mixed result as evidence for two compounding bottlenecks.
First, real-scene object-memory construction introduces substantial upstream noise: crops may be partial, distant, oblique, poorly segmented, or contaminated by background.
Even with best-view selection, some objects lack a clean near-frontal crop, which weakens downstream attribute readout.
Second, some attributes are intrinsically difficult to infer from a single real-scene RGB crop.
Material cues vary with illumination, viewpoint, sensor response, and background contamination, while absolute object size is poorly determined without depth or camera calibration.
This explains why the \texttt{size\_*} probes remain close to random even when semantic or visually distinctive readouts are informative.

\paragraph{Implication.}
This sanity check does not establish deployment-ready real-scene grounding.
Instead, it shows that real-domain failure is not uniform: some learned readouts retain clear signal on real-scene crops, while others remain weak under crop noise and limited real-scene supervision.
We therefore view full real-scene deployment as requiring better object-memory construction---including segmentation, view selection, occlusion handling, and object representation---while preserving the value of attribute readouts as a scene-memory access mechanism.

\end{document}

%% file: color_utils.tex
\definecolor{xblue}{HTML}{4169E1}
\definecolor{xgreen}{HTML}{036C3A}
\definecolor{xpurple}{HTML}{9838B1}
\definecolor{xslategray}{HTML}{70818F}
\definecolor{xorange}{HTML}{FF8C00}
\definecolor{xcyan}{HTML}{06AEEF}
\definecolor{xred}{HTML}{FF0000}
\definecolor{xgray}{HTML}{808080}
\definecolor{xxgreen}{HTML}{009F86}
\definecolor{xsienna}{HTML}{8B4512}
\definecolor{xxpurple}{HTML}{623E99}

\newcommand{\xslategray}[1]{\textcolor{xslategray}{#1}}
